%% file: 0_paper.tex
\documentclass[final]{cvpr}

\usepackage{times}
\usepackage{graphicx}
\usepackage{amsmath}
\usepackage{amssymb}
\usepackage{textcomp}
\usepackage{multirow}
\usepackage[normalem]{ulem}
\usepackage{subcaption}

\usepackage{enumitem}
\usepackage{url}
\usepackage{xcolor}
\usepackage{booktabs} 

\newcommand{\fig}{Fig.\ }
\newcommand{\tab}{Tab.\ }
\newcommand{\sect}{Sec.\ }

\newcommand{\shortcut}{$\mathcal{P}_{\rm shortcut}$ }
\newcommand{\shortcutnows}{$\mathcal{P}_{\rm shortcut}$}

\newcommand{\random}{$\mathcal{P}_{\rm random}$ }
\newcommand{\randomnows}{$\mathcal{P}_{\rm random}$}

\newcommand{\mean}{$\mathcal{P}_{\rm shortcut}^{\rm mean}$ }
\newcommand{\meannows}{$\mathcal{P}_{\rm shortcut}^{\rm mean}$}

\newcommand{\shuffled}{$\mathcal{P}_{\rm shortcut}^{\rm shuffled}$ }

\newcommand{\Ishort}{$\mathcal{I}_{\rm shortcut}$ }
\newcommand{\Ishortnows}{$\mathcal{I}_{\rm shortcut}$}

\newcommand{\Iaug}{$\mathcal{I}_{\rm shortcut}^{\rm aug}$ }
\newcommand{\Iaugnows}{$\mathcal{I}_{\rm shortcut}^{\rm aug}$}

\definecolor{orange}{rgb}{1.0, 0.49, 0.0}

\input{commentCommands} 
\draftfalse
\ifdraft
\usepackage{cancel}
\fi

\usepackage[pagebackref=true,breaklinks=true,colorlinks,bookmarks=false]{hyperref}

\def\cvprPaperID{****} 
\def\confYear{CVPR 2021}

\begin{document}

\title{
Patch Shortcuts: Interpretable Proxy Models Efficiently Find\\ Black-Box Vulnerabilities
}

\author{
    \centerline{{Julia Rosenzweig$^{1}$, Joachim Sicking$^{1, 2}$, Sebastian Houben$^{1}$, Michael Mock$^{1}$, Maram Akila$^{1}$}} \\
    \centerline{$^1$ Fraunhofer IAIS, $^2$ Fraunhofer Center for Machine Learning} \\
    \centerline{\small{\texttt{\{first.last\}@iais.fraunhofer.de}}} \\
}

\maketitle

\begin{abstract}

An important pillar for safe machine learning (ML) is the systematic mitigation of weaknesses in neural networks to afford their deployment in critical applications.
An ubiquitous class of safety risks are \emph{learned shortcuts}, \ie spurious correlations a network exploits for its decisions that have no semantic connection to the actual task.
Networks relying on such shortcuts bear the risk of not generalizing well to unseen inputs.
Explainability methods help to uncover such network vulnerabilities.  
However, many of these techniques are not directly applicable if access to the network is constrained, in so-called black-box setups. 
These setups are prevalent when using third-party ML components.
To address this constraint, we present an approach to detect learned shortcuts using an interpretable-by-design 
network as a proxy to the black-box model of interest.
Leveraging the proxy's guarantees on introspection we automatically extract candidates for learned shortcuts.
Their transferability to the black box is validated in a systematic fashion.
Concretely, as proxy model we choose a BagNet, which bases its decisions purely on local image patches. 
We demonstrate on the autonomous driving dataset A2D2 that extracted \emph{patch shortcuts} significantly influence the black box model.
By efficiently identifying such patch-based vulnerabilities, we contribute to safer ML models.

\end{abstract}

\input{1_intro}
\input{2_related_work}

\input{3_approach}
\input{4_experiments}

\input{5_conclusion}

\paragraph{Acknowledgments}
The research of J.R., M.M. and M.A. was funded by the German Federal Ministry for Economic Affairs and Energy within the project ``KI Absicherung – Safe AI for Automated Driving''. Said authors would like to thank the consortium for the successful cooperation. The work of J.S. was supported by the Fraunhofer Center for Machine Learning within the Fraunhofer Cluster for Cognitive Internet Technologies. The research of S.H. was funded by the Federal Ministry of Education and Research of Germany as part of the competence center for machine learning ML2R (01IS18038B).

{\small
\bibliographystyle{ieee_fullname}
\bibliography{references}
}

\end{document}

%% file: commentCommands.tex
\newif\ifdraft
\usepackage{color}

%% file: 1_intro.tex
\section{Introduction}
\label{sec:intro}

Deep neural networks (DNNs) have demonstrated state-of-the-art performance and good generalization properties on a broad range of tasks. 
However, this generalization ability is not thoroughly understood yet and examples of unexpected failure 
are in stark contrast to the many successful application scenarios. 
These failure cases are 
symptoms of underlying, more fundamental difficulties that come along with recent learning systems.

\begin{figure}[t]
    \centering
    \includegraphics[width=0.9\columnwidth]{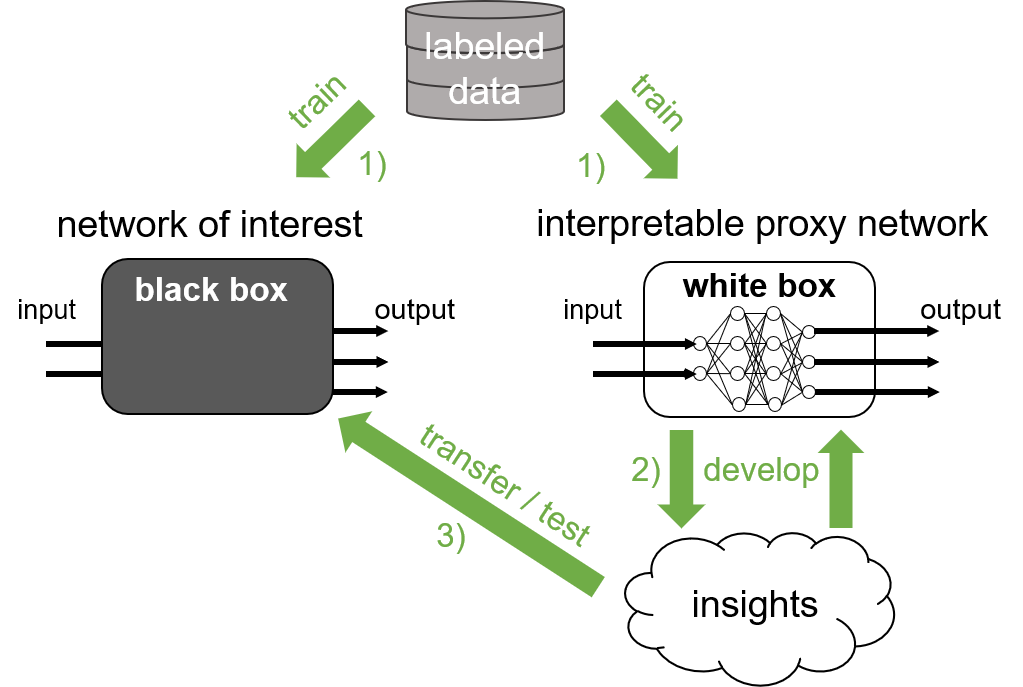}
    \caption{Detecting vulnerabilities of information-constrained black-box models. To avoid cumbersome analyses on the trained black box itself (step 1), we rely on an interpretable proxy model (step 1) that provides a set of so-called (image) patch shortcuts (step 2). Testing these (image) patch shortcuts on the original black box (step 3), we efficiently identify vulnerabilities.}
    \label{fig:schematic_overview_approach}
\end{figure}
Two main challenges for generalization are learned \textit{brittle features} \cite{bug_or_feature} and learned \textit{shortcuts} \cite{shortcut_learning} that are both over-specific for the training (and in some cases even the test) dataset. They differ regarding semantics: brittle feature are mostly considered to be imperceptive statistical artifacts while shortcuts refer to spurious correlations between rather high-level semantic concepts. While both concepts are not fully disjoint, we focus on the more semantic shortcuts here.

To illustrate how networks exploit correlations for their decisions that have no semantic connection to the actual task, we outline three examples: 
first, the poster child of shortcut learning. Images (\eg showing horses) that are in fact classified due to imprinted watermarks or copyright tags \cite{horse_watermark}.
Second and more critically, models to detect Covid-19 in \eg chest radiographs that rely on non-medical shortcuts (in particular outside the lungs) that are moreover hospital-specific \cite{Covid19_shortcuts}. 
Lastly, in the field of autonomous driving shown here this could \eg be a patch showing parts of windows in a building which the model erroneously identifies as parts of ``car'', see left panel of \fig \ref{fig:approach_exampl}.

Various explainability techniques were put forward to uncover such network vulnerabilities \cite{GradCAM, interpretability_review, visualizing_DNNs}. 

However, information-constrained black-box setups limit the applicability of most of them.
At the same time, handling these black boxes gains importance as ML models are increasingly used in commercial contexts ranging from ML-as-a-service to ML appliances from technical providers. 
For, \eg, auditors and regulators with limited mandates such set-ups can pose additional challenges.

In this work, we propose an approach to systematically and efficiently detect learned shortcuts of a black-box classification model.
Our approach, as depicted in Fig.~\ref{fig:schematic_overview_approach}, consists of three steps:  
in a first step, we train a white-box proxy model, namely BagNet that is interpretable-by-design. 
Next, we make use of its ``locality'' guarantee and systematically extract patch shortcuts, \ie semantic vulnerabilities, of the proxy network. 
In the final step, we evaluate if, and to which extent, the identified patch shortcuts transfer to the black-box network of interest. 

The remainder of the paper is organized as follows: 
First, we present related work on shortcut learning, interpretable-by-design architectures, approaches using proxy models and black-box vulnerabilities in \sect \ref{sec:related_work}. 
Next, we outline how to find and test patch shortcuts using an interpretable proxy network in \sect \ref{sec:patch_shortcut}. 
We instantiate this approach for a binary classification network in the automotive domain and conduct different ablation analyses to validate this patch-shortcut-based testing in \sect \ref{sec:experiments}. 
An outlook in \sect \ref{sec:outlook} concludes the paper.

%% file: 2_related_work.tex
\section{Related work}
\label{sec:related_work}

Our approach resides at the interface of several research directions, connections to which we detail here. We focus on shortcut learning and interpretability-by-design but also (patch-based) augmentations, the question of transferability of results, and adversarial vulnerabilities, specifically for black boxes, as the success of our shortcut patches is measured by decreased (black-box) DNN performance.

\paragraph{Shortcut learning}
In our work, we concentrate on finding vulnerabilities in the form of \textit{shortcuts} in images.
For a more general overview on shortcut learning see \eg \cite{shortcut_learning}.
Recent work~\cite{Covid19_shortcuts, xray_shortcuts} analyzes the aforementioned shortcuts in the domain of medical imaging and emphasize the necessity to validate the reliability of ML models. 

Close to our idea is the work by Shetty \etal~\cite{context_effects}, in which the role of context for classification and segmentation is inspected and, in doing so, also the question of whether meaningful concepts were learned is addressed.
However, they remove (human understandable) objects (\eg cropping out the cars based on their segmentation mask to see how this affects sidewalk segmentation) whereas we (re-)insert (not necessarily human understandable) patches from the dataset into the image. 
Rosenfeld \etal~\cite{elephant_in_the_room} either insert some random, trained objects or remove objects and reinsert them into the images at a different location to see how this affects object detection. 
They observe that their ``transplantations'' have non-local effects to objects ``far away''.\footnote{This
susceptibility to position is also observed by Kayhan \etal ~\cite{translation_invariance}.} 
We conduct a similar analysis systematically evaluating the inserted patches' sensitivity to chosen positions.
In addition, we use an informed approach of selecting the parts which are to be inserted and in particular, we do not crop whole objects but only patches.

\paragraph{Interpretability-by-design and surrogate models}
Particularly related to our approach are \textit{interpretable-by-design networks}  such as invertible networks, in which information is processed in a bijective way ~\cite{irevnet, iresnet}. 
However, due to the special way of processing information, the applicability of these invertible networks is limited.
Ghorbani \etal~\cite{concept_based_explanations} extract semantic visual concepts important for the decision making.
While their approach involves a segmentation that uses global information, BagNets aggregate purely local information in a linear fashion. 
Using human-understandable concepts, the ProtoPNet \cite{prototypical_part_net} is based on comparisons of the current input to prototypical class parts it learned. 
The network-in-network architecture \cite{network_in_network} enables discrimination between local patches in the receptive field.

The idea of using \textit{surrogates} for explaining black-box decisions is used in \eg  \cite{distill_soft_decision_tree, symbolic_metamodel, distill_and_compare, LIME}.
Mohseni \etal~\cite{predicting_failure_modes_saliency} use a student model to predict the teacher's failure modes from the teacher's saliency maps in a white-box setup. While there are some heatmap methods for the black-box setup, see \eg \cite{RISE, bb_explainability}, many common approaches rely on backward gradients and are thus not easily applicable as interpretability methods for our black-box setup.
Often distillation \cite{distillation, born_again} is employed to train a surrogate model.

\paragraph{Patch-based augmentations} 
Many approaches take small rectangular pieces of an image, so-called patches, as their starting point: In \emph{image quilting} \cite{quilting}, small pieces of training images are recomposed to resemble inputs presented at inference, thus
enabling texture synthesis and texture transfer for whole objects and images. \emph{Patch-based augmentations} (\eg \cite{augmix, cutout, autoaugment}) are frequently used for model training aiming at improving performance or robustness. 

\paragraph{Adversarial vulnerabilities}
The \textit{black-box} setup with access to solely input-output pairs is common in the context of \textit{adversarial attacks}, see \eg~\cite{bb_attacks}. 
Many approaches use \textit{surrogate models} and \textit{transferability}-based methods in which attacks are crafted on a white box and successfully transferred
to the target black box, see \eg \cite{transferability_bb, transferability}.\footnote{Naseer \etal~\cite{transferability} show that 
this transfer is not only possible across networks but also across domains.} 
Particularly, \textit{Jacobian-based saliency map attacks} \cite{limitations_of_dl, maximal_JSMA, probabilistic_JSMA} are a related research direction. 
They employ a saliency map based on forward gradients to find ``influential'' pixels/features they seek to manipulate.
We do not aim at pixel or feature manipulations and thus neither change nor optimize identified patch shortcuts.
This key fact distinguishes our approach from most works on adversarial attacks.
Additionally, we want to mention \textit{semantic attacks} in which images are modified along human-understandable, semantic dimensions. For example,  hue and saturation or colorization and texture of images, respectively, are randomly perturbed while keeping the semantic concept fixed \cite{semantic_adv_ex, Unrestricted_adv_examples}.
Jacobsen \etal~\cite{excessive_invariance} show that networks are often invariant to concepts that are relevant to the task and too sensitive to irrelevant ones. 

Delineated as a reinforcement learning problem, Yang \etal~\cite{PatchAttack} insert patches with textures into images to fool classifiers, an approach  applicable to the black-box setting. 
We state two key differences: First, instead of learning or optimizing texture, we insert original image patches, hence, by design remaining in the domain of natural image patches.\footnote{See also Hendrycks \etal~\cite{natural_adv_example}, who gather natural images that confuse models in a dataset, and the real-world attack using stickers \cite{stickers_on_stop_signs}, which, however, does not allow for systematic analysis of failure modes.} 
Second, as mentioned above, we do not optimize for the size or the position of the patch.
Other work in the field of attacks focuses on \textit{patch-based attacks}, \eg~\cite{adv_patch, fooling_person_detectors_patch, physical_adv_patch, square_attack, dpatch}, in which, however, patches are optimized.

%% file: 3_approach.tex
\section{Patch shortcuts}
\label{sec:patch_shortcut}

In this section, we detail the conceptual approach of finding patch shortcuts for a given trained black-box classification network. We assume access to the training data of the black box and knowledge of the task it was trained for. Apart from this, we can only  query the network output (oracle access).

The structure of this section follows our proposed three-step approach shown in \fig \ref{fig:schematic_overview_approach}. We describe details regarding 
training the proxy network, automatically developing insights on it, and finally systematically testing these on the black box in the following subsections. 


\subsection{Training the interpretable proxy network}
\label{sec:approach_step1}

\begin{figure*}[hbt!]
    \centering
    \includegraphics[width=\textwidth]{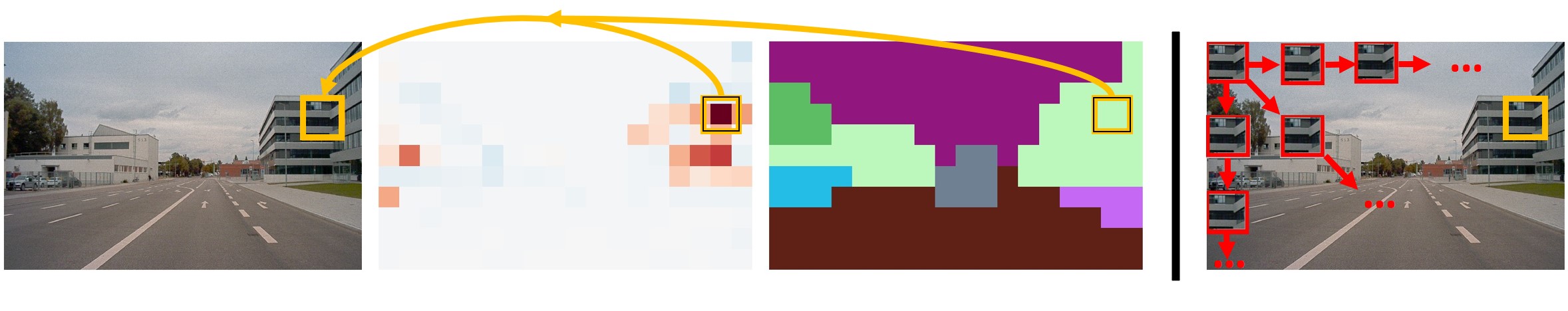}
    \caption{Identifying (first three panels) and testing (last panel) patch shortcuts. 
    A patch is a BagNet patch shortcut (orange box in first panel) for class ``car'' if two conditions are fulfilled: 
    First, it is highly relevant for BagNet's ``car'' prediction (dark red on heatmap in second panel),
    second, its semantic class is not ``car'' (light green segment in the third panel). To systematically test such a BagNet patch shortcut, we create a dedicated testing dataset (last panel).  
    }
    \label{fig:approach_exampl}
\end{figure*}

\paragraph{Dataset requirements}
Typically, annotations of a dataset are designed and used to train a specific task, \eg class labels or segmentation masks. This, however, implies that all parts of the annotation are encoded in the network and cannot easily be used to identify shortcuts learned by the DNN. To investigate shortcut learning, we deliberately use an under-specified toy task, \ie learn a binary classification task for a chosen class of interest (\eg presence of cars) on a public dataset for semantic segmentation.
In doing so, the segmentation mask serves as \textit{meta annotation} to identify possible shortcuts (\ie relevant image regions not occupied by cars) \cite{underspecification}. 
Although the overall approach can be applied for different tasks with the appropriate choice of interpretable network (see \sect \ref{sec:outlook}), we detail the concept for the case of image classification using the so-called BagNet trained on the aforementioned binary classification task as interpretable-by-design network.  

\paragraph{BagNets} 
BagNets are based on the ResNet-50 architecture~\cite{resnet} with some of the $3\times3$ convolutions replaced with $1\times1$ convolutions.
This results in a strictly smaller receptive field than usual ResNet architectures possess.\footnote{BagNet architectures with receptive fields of $9\times9, 17\times17$ and $33\times33$ pixels, respectively, have been proposed in literature~\cite{bagnet}.}
BagNets perform an individual linear weighting of each ``pixel'' in the last feature map, effectively yielding patch-wise classification logits for the input image.
Here, it is important to note that BagNets only aggregate information from image patches with the size of the receptive field, thus relying on truly local features and not aggregating or mixing evidence from across the entire image.\footnote{Please note that the patch regions are partly overlapping due to the used stride in the BagNet.} 
These patch-wise logit maps serve as internal, strictly feed-forward heatmaps (one for each class), using solely local patch information, which is computed in a single forward pass. See the heatmap on the left of \fig \ref{fig:approach_exampl} as an example for a ``car'' heatmap. 
Averaging the evidence from these heatmaps yields the final logit output of the BagNet for each class.
Please note that to generate the described patch-logit heatmaps, we need to exchange two commutative layers in the BagNet definition so that the fully connected layer is applied before 2D average pooling. This allows us to save the representation after the fully connected layer as our heatmap.

\paragraph{Using BagNets as interpretable local-feature proxies} We choose BagNets as interpretable proxies over producing \eg occlusion-based forward heatmaps on the ResNet directly mainly for two reasons:
First, this approach is much more efficient since a BagNet heatmap is produced in a single forward pass whereas a forward heatmap for a black box would require multiple rounds of inference with systematically shifted occlusions. 
Second and more important, we want to exploit the locality property of BagNets as they do not rely on global, long-distance features across the image but aggregate evidence by averaging local patches, see \cite{bagnet}.
This enables explanations of model outputs in terms of each individual image patch irrespective of its position in the image.

\subsection{Finding patch shortcuts}
\label{sec:approach_step2}

As a second step we use the trained BagNet to find semantic vulnerabilities. 
That is, for our selected class of interest, we identify image patches that are highly predictive 
for one class yet actually stem from a different class, \ie semantic concepts that correlate with the class of choice although not related semantically. 
We  detect these highly relevant patches in the following way: 
We perform inference on the whole test set using our adapted version of BagNet. 
We only consider those patches whose  logit for the heatmap of the targeted class is above the $99\%$ quantile $q_{\rm logit}^{0.99}$ of 
logit values over all patches from the dataset
and, thus, highly predictive of our class of interest.
Next, for each obtained patch, we consult the corresponding part of the segmentation mask to verify whether it contains parts of the targeted class?.\footnote{Classification towards a class could, \eg, be caused by small parts of an object of said class overlapping with the receptive field, a scenario we want to avoid with respect to shortcuts.}
If not, we identified a patch \textit{shortcut} for this class whose prediction only correlates with the chosen class while holding no direct semantic relation. 
This procedure is depicted using an example on the left of Fig.~\ref{fig:approach_exampl}.

\subsection{Testing transferability to the black box}
\label{sec:approach_step3}

The last step of our approach aims to evaluate to which extent the identified BagNet shortcuts also constitute shortcuts, and thus semantic vulnerabilities, of the black-box network. 
For that, we propose the following procedure of constructing a testing image set: Using the set
of images containing the patch shortcuts identified in \sect \ref{sec:approach_step2}, we 
consider the subset of images which are correctly classified by either the black-box or BagNet network as \textit{not} belonging to the class of interest, denoting this set by \Ishortnows. The respective set of patch shortcuts extracted from the images in \Ishort is denoted by \shortcutnows.
Then, we automatically copy each patch shortcut from \shortcut 
and re-insert it into the same image 
but at a new position.
We then provide this manipulated image as input to the black-box network.
We consider an insertion \textit{successful} if it changes the prediction of the network to a misclassification.

To account for possible position-sensitivity of the black-box network, as observed in other work, \eg~\cite{translation_invariance, elephant_in_the_room}, we insert each identified BagNet patch shortcut from \shortcut in a grid-based manner into many distinct positions of the original image. More concretely, we insert it into every position that corresponds to exactly one patch logit value (``pixel'') of the BagNet heatmap. 
Note that we insert only a single patch at a time but at varying positions, resulting in a total number of manipulated images equal to the amount of ``pixels'' in the BagNet heatmap.\footnote{Please note that one of these positions necessarily corresponds to the original image as the patch is replaced with itself.} The testing image set obtained that way is referred to as \Iaugnows.
This procedure is depicted using an example on the right of \fig \ref{fig:approach_exampl}. 
Finally, we statistically evaluate to what extent the black box is susceptible to the BagNet shortcut patches by conducting inference on \Iaug and thereby analyzing how often each patch leads to misclassifications if inserted into the image at all possible positions.

%% file: 4_experiments.tex
\section{Experiments}
\label{sec:experiments}
Having outlined our approach to finding and testing black-box 
patch shortcuts, 
we now instantiate and evaluate it for shortcuts for the class ``car'' deploying a classification network from the automotive domain. 
The structure of the section follows the steps introduced in \sect \ref{sec:patch_shortcut}:
After detailing the dataset and training procedures in \sect \ref{sec:experimnts_step1}, we generate the shortcuts in \ref{sec:experiments_step2}
and systematically test these on the black-box network in \sect \ref{sec:experiments_step3}.
To judge the effectiveness of our approach, baseline as well as further ablation studies are conducted in \sect \ref{sec:experiments_baseline} and \ref{sec:experiments_ablation}. 

\subsection{Training the interpretable proxy network}
\label{sec:experimnts_step1}

We first describe the custom dataset on which all experiments in this section are performed. 
Subsequently, the training configuration for the interpretable white-box model (and also the black-box model) is presented.

\paragraph{Dataset}
A2D2 \cite{a2d2} is a sequence-based traffic-scene dataset containing $41{,}277$ images
that provides (among others) semantic segmentation ground truths.
However, we do not intend to segment input images but to classify them either as ``car'' if one or several cars are displayed or as ``no-car'' otherwise. 
More specifically, images that feature at least $ 2\%$ ``car'' pixels belong to class ``car'' and images containing  $ 0.3\%$  ``car'' pixels or fewer are counted as ``no-car''.
All other images are discarded.\footnote{In particular, this means that an image labeled as \eg ``no-car'' can still contain very few car pixels, which are however negligible \wrt the total image area.}
We refer to the resulting dataset as \textit{binary-classification A2D2} or just \textit{binary A2D2}.
Since binary-classification A2D2 does not require the full segmentation ground truth, it allows us to use this ground truth as \textit{pixel-wise meta-annotation} instead.
To prevent both data leakage and domain shift between train and test data,
we split each sequence from the dataset into three equally sized sub-sequences
and apply a random $80$:$20$ train-test split on sub-sequence level.
For training and evaluation, the images are resized to $100\times160$ pixels and normalized.

\paragraph{Networks and training}
As our black-box model, we choose a ResNet-50 that is trained on binary-classification A2D2 for $150$ epochs with a batch size of $128$ 
using the Adam optimizer \cite{adam} with an initial learning rate of $0.001$ and a binary cross-entropy loss.
Unsurprisingly, the resulting model yields 
an almost perfect performance on the test set ($\rm{acc} = 0.9748$, $F_1 = 0.9556$). We again point out that our analyses target the question of \textit{how} these classifications are made and an over-simplified task provides a reasonable setup for such a study. 
In the following, we refer to this trained ResNet as black box (BB) network and do not make use of any network-internal properties. 

As white-box proxy network, we employ an interpretable-by-design BagNet with a receptive field of $17\times 17$.\footnote{We use the BagNet architecture provided here: \url{https://github.com/wielandbrendel/bag-of-local-features-models}} 
Its training configuration does not differ from the one above, except for a smaller batch size of $64$. 
This BagNet reaches a test 
performance of $\rm{acc}=0.9695$ and $F_1 = 0.9455$.

\subsection{Finding patch shortcuts}
\label{sec:experiments_step2}

\begin{figure}[h]
    \centering
    \subfloat[A BagNet patch shortcut showing trash cans.]
    {\includegraphics[width=0.99\columnwidth]{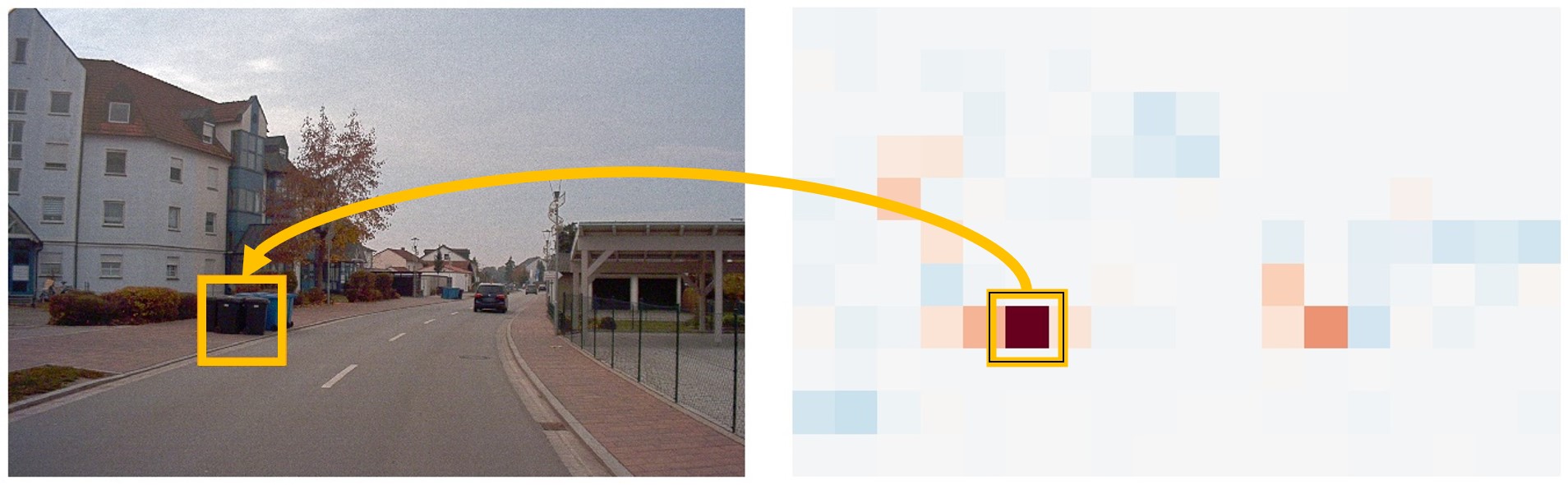}}
    
    \subfloat[A BagNet patch shortcut showing a curb.]
    {\includegraphics[width=0.99\columnwidth]{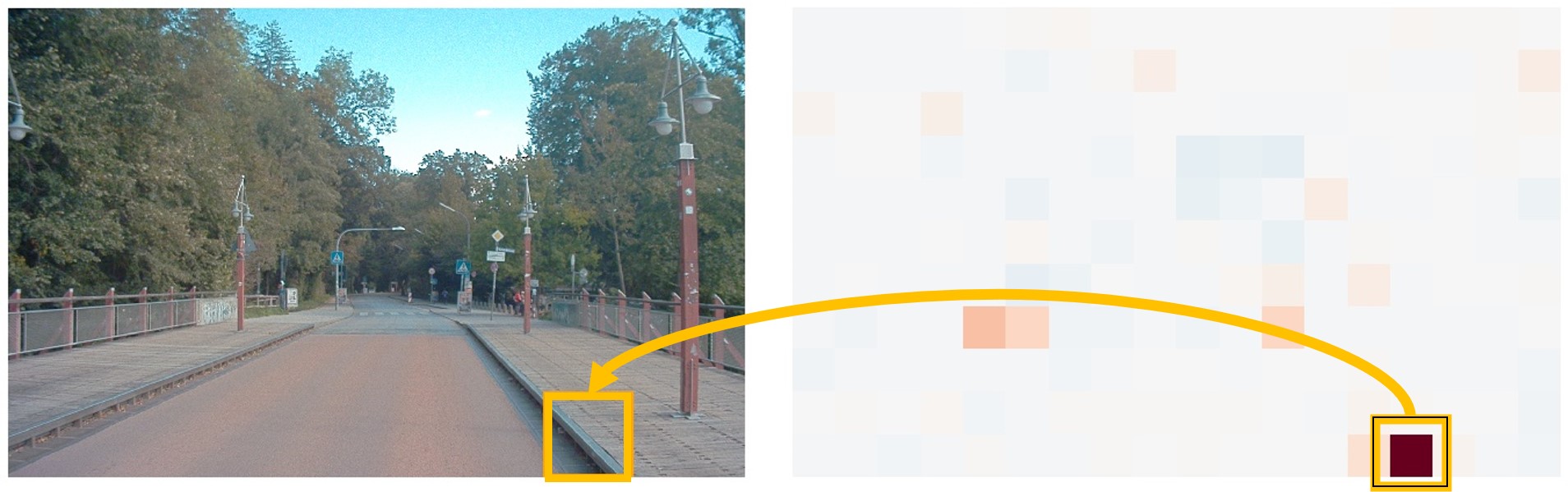}}

    \caption{Identified BagNet patch shortcuts (orange frames). We combine the semantic pixel-wise annotations and BagNet's intrinsic forward heatmap (rhs: light-blue means low evidence for ``car'', dark red means high evidence for ``car'') to identify shortcuts (see lhs), \ie ``no-car'' patches that BagNet correlates with class ``car'', see text for details.
    }
    \label{fig:example_patches}
\end{figure}

Using the trained BagNet, our local-feature white-box proxy, we follow the procedure described in \sect \ref{sec:approach_step2} to find patch shortcuts \shortcutnows: The BagNet's forward heatmap\footnote{The heatmap is of size $11\times 18$ pixels in our case.} and the semantic segmentation ground truth mask are compared  and the combined selection criterion is applied (see \sect \ref{sec:approach_step2}). 
To investigate the semantic origin of patches we use a down-sampled version of the segmentation mask, compare \fig\ref{fig:approach_exampl}.\footnote{If the part of the segmentation mask corresponding to the patch contains the class in question, \ie ``car'' or related classes, we count the down-sampled result as ``car''. Otherwise a majority voting among the other classes present in the patch is used.}
Two examples of such patch shortcuts for class ``car'' and their corresponding BagNet heatmap are displayed in \fig \ref{fig:example_patches}.

We observe that many shortcut patches belong to the semantic classes of ``nature object'', ``building'' or ``obstacle/trash''. 
Moreover, edges seem to be common features of patch shortcuts (see \eg the bottom row of \fig \ref{fig:example_patches}).

\subsection{Testing transferability to the black box} 
\label{sec:experiments_step3}

We follow the procedure described in \sect \ref{sec:approach_step3} to evaluate if, and to what extent, the identified BagNet shortcut patches in \shortcut are important for the black-box classifier. 
To enable systematic testing, we create the separate testing dataset \Iaug
for each identified patch shortcut from \shortcut by duplicating the respective image (from \Ishortnows) and copying the identified patch shortcut to different positions (see last panel of \fig \ref{fig:approach_exampl}).
We evaluate the black-box model on both the undistorted image dataset \Ishort (that only contains the naturally occurring shortcuts) and the patch-shortcut-augmented dataset \Iaug and compare their (normalized) confusion matrices (see bottom row of \tab \ref{tab:confusion_mat_resnet_bagnet}). 
For comparison, we also report the respective results when applying the BagNet instead of the black-box model (see top row of \tab \ref{tab:confusion_mat_resnet_bagnet}). 
For both networks, a shift from true negative (TN) to false positive (FP) can be observed after inserting the patch shortcuts.
As expected, true positive (TP) images are mostly unaffected by patch-shortcut insertions, see the virtually unchanged second and fourth row of \tab \ref{tab:confusion_mat_resnet_bagnet}.
This preliminaryly shows that the identified BagNet shortcuts also constitute shortcuts for the black box as they are able to provoke misclassifications.

\begin{table}[ht!] 

\centering
\par\bigskip
\setlength\tabcolsep{6pt} 
\begin{tabular}{@{}cc cc cc c@{}}

\multicolumn{1}{c}{} &\multicolumn{1}{c}{} &\multicolumn{1}{c}{} &\multicolumn{4}{c}{Predicted Class} \\ 
\cmidrule(lr){4-7}
\multicolumn{1}{c}{} &\multicolumn{1}{c}{} &\multicolumn{1}{c}{} &\multicolumn{2}{c}{\Ishort} &\multicolumn{2}{c}{\Iaug} \\ 
\cmidrule(lr){4-5} \cmidrule(lr){6-7}
\multicolumn{1}{c}{} & 
\multicolumn{1}{c}{} & 
\multicolumn{1}{c}{} & 
\multicolumn{1}{c}{no-car} & 
\multicolumn{1}{c}{car} &
\multicolumn{1}{c}{no-car} & 
\multicolumn{1}{c}{car} \\
\cline{3-7}

\addlinespace[0.5em]
\multirow[c]{4}{*}[-1pt]{\rotatebox[origin=tr]{90}{Actual Class}} & 
\multirow[c]{2}{*}[3pt]{\rotatebox[origin=tr]{90}{BagNet}}
& no-car & $0.63$ & $0.37$ & $0.35$ & $ 0.65 $ \\
\addlinespace[0.25em]
& & car & $0.01$ & $0.99$ & $0.00$ & $1.00$ \\ 
\addlinespace[0.25em]

\cline{3-7}

\addlinespace[0.5em]
 & \multirow[c]{2}{*}{\rotatebox[origin=tr]{90}{BB}}
& no-car & $0.90$ & $0.10$ & $0.80$ & $ 0.20 $ \\
\addlinespace[0.25em]
& & car & $0.00$ & $1.00$ & $0.00$ & $1.00$ \\ 
\addlinespace[0.25em]

\cline{3-7}
\addlinespace[0.5em]
\end{tabular}
\caption{Normalized confusion matrices of BagNet (first two rows) and the black box (BB, last two rows) before (first two columns) and after (last two columns) insertion of BagNet patch shortcuts. Note that each identified BagNet patch shortcut is inserted into $11 \times 18 = 198$ distinct positions in the image. Hence, one insertion position corresponds to exactly the original image.
}
\label{tab:confusion_mat_resnet_bagnet}
\end{table}

\subsection{Random image patches as baseline}
\label{sec:experiments_baseline}
To check the soundness of our approach,
we compare the set of patch shortcuts, \shortcutnows, with a \textit{random} ``no-car'' patch set, denoted as \randomnows, that is obtained using random ``no-car'' patches in the \Ishort images from the $50\%$ logit quantile.\footnote{Please note that we use only such images that contain patches from both logit quantiles.}  
In total, we consider image patches from $96$ distinct images for this analysis.

For both BagNet and the black-box model, we report the mean and median numbers of successful patch insertions for patches in \shortcut (first column of \tab \ref{tab:experimental_results}) and patches in \random~(second column of \tab \ref{tab:experimental_results}). 
We find patches from \shortcut to be more successful by a big margin.\footnote{As expected, this effect is even stronger for the BagNet since it was used for patch selection. Please note that due to the overlapping receptive field of the BagNet, every insertion also slightly manipulates patch logits of neighboring patches in the heatmap by introducing edges and thus artifacts in neighboring patches. Thus, not all the insertions can be expected to be (equally) successful on the BagNet.}
Note that all example images in this work display patches from \shortcut (in orange frames) that are among the most successful ones on the black box.

\begin{table}[ht!] 
\centering

\setlength\tabcolsep{3pt} 
\begin{tabular}{@{}cc cc cc@{}} 

\multicolumn{2}{c}{} & \multicolumn{4}{c}{Patch Set} \\ 
\cmidrule(lr){3-6}
\addlinespace[0.25em]
\multicolumn{1}{c}{} & 
\multicolumn{1}{c}{} & 
\multicolumn{1}{c}{\shortcut} & 
\multicolumn{1}{c}{\random} &
\multicolumn{1}{c}{\mean} & 
\multicolumn{1}{c}{\shuffled} \\
\addlinespace[0.25em]
\cline{2-6}

\addlinespace[0.5em]
\multirow[c]{2}{*}{\rotatebox[origin=tr]{90}{BagNet}}
& mean ($\uparrow$) & $\mathbf{67.42}$ & $10.65$ & $16.84$ & $12.24$ \\
\addlinespace[0.25em]
& median ($\uparrow$) & $\mathbf{28.50}$ & $1.00$ & $3.50$ & $1.00$ \\
\addlinespace[0.5em]

\cline{2-6}

\addlinespace[0.5em]
\multirow[c]{2}{*}{\rotatebox[origin=tr]{90}{BB}}
& mean ($\uparrow$)  & $\mathbf{20.59}$ & $11.86$ & $7.98$ & $5.27$ \\
\addlinespace[0.25em]
& median ($\uparrow$) & $\mathbf{17.00}$ & $7.00$ & $1.67$ & $1.25$ \\
\addlinespace[0.5em]

\cline{2-6}
\end{tabular}
\caption{Mean and median number of successful patch insertions for the BagNet and the black box (BB) per image (higher is better as this means that the network is more susceptible to the patch). We report the results from our main and baseline experiment (first two columns) as well as from the further ablation study (last two columns, see \sect \ref{sec:experiments_ablation} for details).  Note that each patch is inserted into $11\times18 = 198$ positions in the image. }
\label{tab:experimental_results}
\end{table}

\begin{figure}[hbt!]
    \centering
        \subfloat
        {\includegraphics[trim = 0 0 68 0, clip, width=0.465\columnwidth]{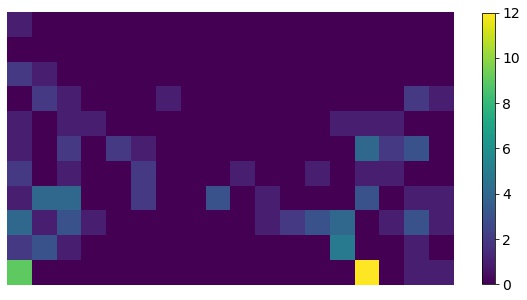}}
        \subfloat
        {\includegraphics[width=0.535\columnwidth]{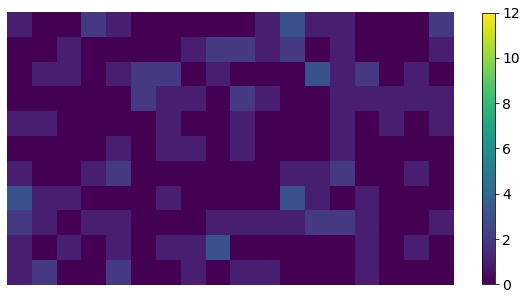}}
        
        \subfloat
        {\includegraphics[trim = 0 0 68 0, clip, width=0.465\columnwidth]{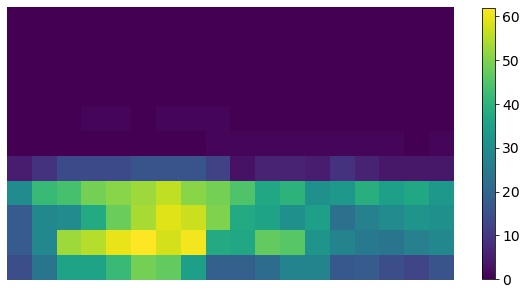}}
        \subfloat
        {\includegraphics[width=0.535\columnwidth]{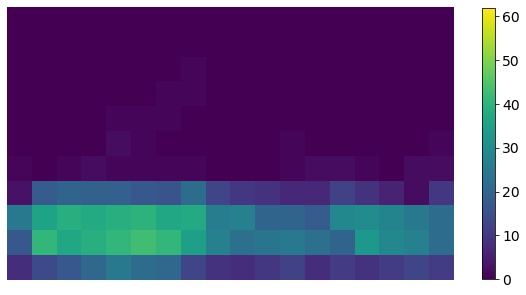}}

    \caption{Analysis of the patch sets \shortcut (left panel) and \random (right panel) regarding position. The heatmaps (blue is 
    rare, yellow is 
    frequent) on the top visualize the origins of the patches in the respective patch set and the ones on the bottom show the positions where patch insertions switched the respective black-box prediction from ``no-car'' to ``car''.
    In this regard, patches from \shortcut are clearly more effective than random ones from \random\ as they can be placed more ``freely'', \ie in more possible positions, in order to provoke misclassification.
    }
    \label{fig:origins_positions_q99_q50}
\end{figure}

Next, we inspect the origin of the patches in \random~and \shortcut and study the \textit{position sensitivity} of the black-box network in more detail (see Fig.~\ref{fig:origins_positions_q99_q50}). 

Regarding the patch origin, we observe that the informed and more successful patches in \shortcut stem from two triangle-shaped regions of the input images,
indicating that semantic concepts to the sides of the road contain shortcuts. 

To verify our procedure, we also show the origins of the patches from \random which exhibit uniform distribution as expected.

For the black-box model, we observe a high sensitivity to patches inserted in the bottom part of the image (corresponding to common locations of cars on the road) and almost no effect when patches are inserted 
in the upper part of the image, \ie in the region above the street level. 
Moreover, the left hand side of the bottom part (corresponding to the oncoming lane) shows the highest sensitivity to patch insertions. 
Overall, we observe that patches in \shortcut are less susceptible to the insertion location compared to patches from \randomnows.
This lends credence to the fact that our patches in \shortcut carry relevant shortcut information the black box is sensitive to.

\subsection{Further ablation analyses}
\label{sec:experiments_ablation}

 \begin{figure}[hbt!]
     \centering
     \includegraphics[width=0.9\columnwidth]{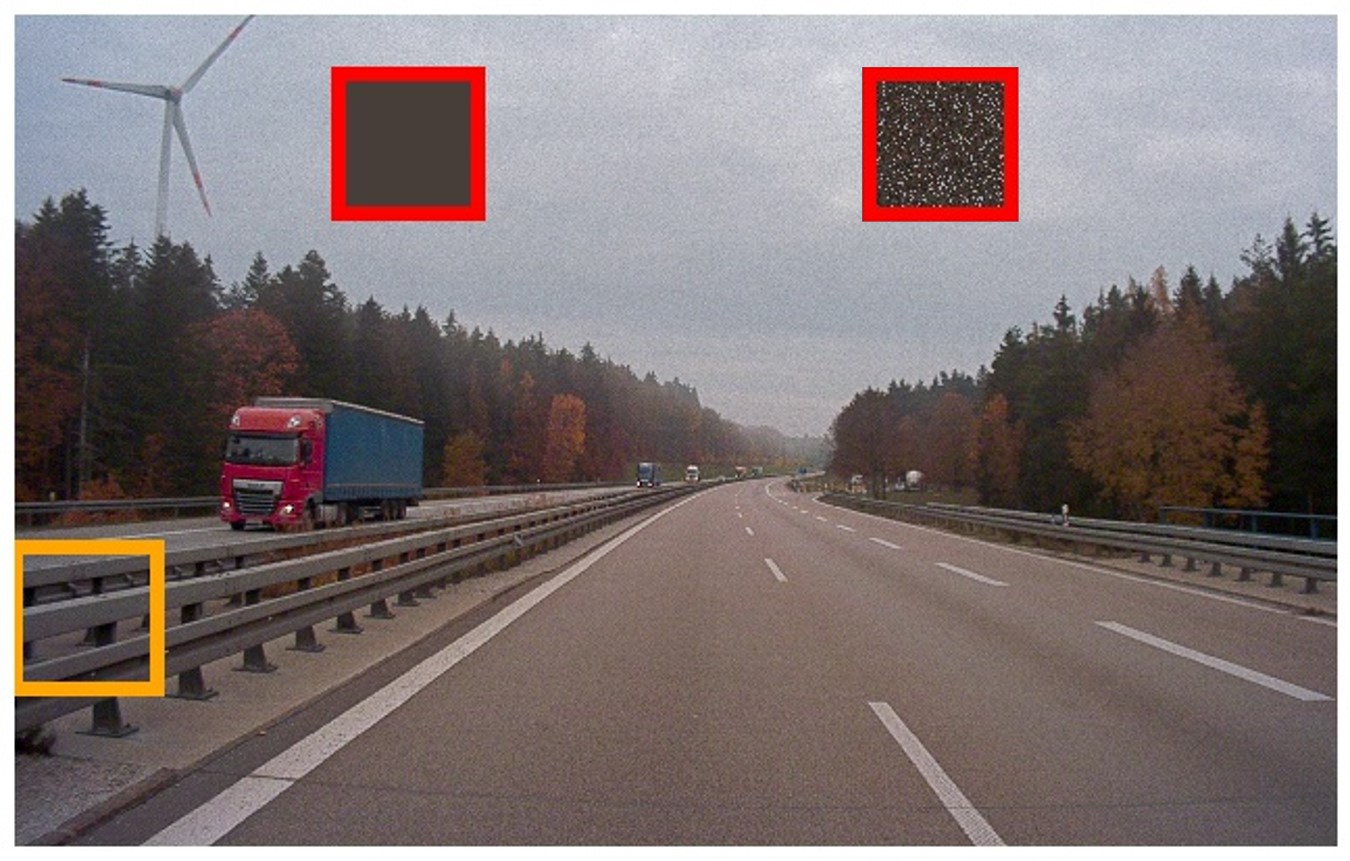}
     \caption{Illustration of a patch shortcut (orange frame) that is averaged (left red frame) or shuffled (right red frame) before being inserted into another part of the image. Different from this visualization, we never insert more than one patch into an image.} 
    \label{fig:ablation_patches}
 \end{figure}

In the above experiments, we find insertions of patches from \shortcut  to be more effective compared to those from \randomnows.
However, all these patch insertions introduce edge effects \wrt the 
surrounding image information and thus artifacts that might influence the behavior of the network. 
To better understand the impact of such artifacts, two ablation experiments (based on the patches in \shortcutnows) are performed:
We either shuffle the pixels inside the shortcut patch before insertion or replace them by their mean values, denoting the obtained patch sets by \shuffled and \meannows, respectively.
Shuffling the patch pixels removes the spatial relations between the pixels while keeping the color distribution unchanged. 
Replacing all pixels by the average color collapses this 
distribution onto its mean value.
Both variants erase most of the semantic information that the original patch carried, see Fig.~\ref{fig:ablation_patches} for an example of both. 
This allows us to disentangle the effect of edge artifacts and of semantic concepts. 
The statistical analysis is performed as in \sect \ref{sec:experiments_step3} above and
the results are shown in the last two columns of \tab \ref{tab:experimental_results}. 
We find the mean and median numbers of successful patch insertions to drop significantly---even if compared to patches from \randomnows. 
The outcomes provide evidence for a ``base'' effect that can be attributed to edges and further artifacts since both ablations lead to a small number of successful patch insertions. 
This base effect, however, is minor compared to the effect of semantically intact patch shortcuts, thus, stressing the semantic meaningfulness and effectiveness of our approach.


%% file: 5_conclusion.tex
\section{Conclusion}
\label{sec:outlook}


We introduced an approach to identify learned shortcuts of a black-box network by analyzing a white-box proxy network, namely an interpretable-by-design BagNet that builds on local feature statistics. 
The patch shortcut candidates extracted via BagNet are transferred to and tested on the black-box model. 
The empirical evaluation on the binary-classification A2D2 dataset demonstrated the efficacy of our approach. 
Detected BagNet patch shortcuts, if tested on the black box, lead to significantly more misclassifications of the considered black-box network than, \eg, random insertions.
Hence, they enable us to efficiently find vulnerabilities of the black box. 
An ablation study demonstrated that only a smaller fraction of the observed effects 
can be explained by
edge artifacts. 
Most of it can be attributed to the semantics of the patch.


The employed coupling between BagNet and black box is relatively loose:
Both networks are trained for the same task on the same dataset but, apart from this, share no information. 
We therefore expect more direct couplings, \eg teacher-student approaches \cite{distillation}, to show even higher transferability rates. 
Further investigating how shortcut transfer depends on the chosen coupling technique seems promising.


Employing BagNet, our approach leverages the ``locality'' guarantees provided by this specific interpretable-by-design model. 
However, there are other classes of interpretable models such as invertible architectures, \eg iRevNet \cite{irevnet}, providing different guarantees and thereby offering alternative means for model assessment.
It might further be possible to lift the need for fine-grained meta-annotations (in our case the pixel-wise semantic ``car'' or ``no-car'' information): 
Clustering the image patches that are crucial for BagNet decisions would enable a human-in-the-loop to readily detect predominant shortcut concepts.


Transferring the approach of a patch-based proxy model from images to other types of unstructured data, \eg audio, video or text seems feasible. 
The concept of ``image patch'' then translates to short audio snippet or frequency band, volumetric cube or chunk of words.


Having instantiated our approach on a toy task, we emphasize that shortcut learning is by no means limited to such simple setups. 
It is a problem of more general nature \cite{shortcut_learning} and shortcuts are an ``ubiquitous'' property of ML, affecting it for tasks of various complexity. 
As part of future work, one could extend the analysis to other datasets, more complex tasks and various kinds of black-box models.
A systematic analysis of learned shortcuts, as made possible with the presented approach, contributes to safe ML by an early discovery of potential weak spots and failures. Further on, such insight opens the possibility for mitigation. 
Similar to \eg adversarial training, shortcut patches could be used to augment and robustify training procedures. Using the image augmentation presented here not for testing but to generate new images, one could increase the prevalence of identified shortcuts within the dataset. To avoid misclassification, a network trained on this enhanced data would have to be more robust with respect to those shortcuts. Furthermore, a \emph{shortcut exploitation score} reflecting the vulnerability of a given DNN could be used as secondary metric for model comparison.